\documentclass[pmlr]{jmlr}% new name PMLR (Proceedings of Machine Learning)

 \usepackage{booktabs}
 \usepackage{algorithm}
 \usepackage{algorithmic}

\RequirePackage{graphicx}
 \usepackage{booktabs}
\usepackage{longtable}% for long tables
\usepackage{comment}
\usepackage{enumitem}
\usepackage{graphicx}
\usepackage{subcaption}
\usepackage{graphicx}
\usepackage{comment}
\usepackage{svg}
\usepackage{algorithm}
\usepackage{graphicx}
\usepackage{subcaption}
\usepackage{amsmath}
\DeclareMathOperator*{\argmin}{arg\,min}

 \usepackage{comment}
\usepackage{longtable}% for long tables
\makeatletter
\def\set@curr@file#1{\def\@curr@file{#1}} %temp workaround for 2019 latex release
\makeatother
\usepackage[load-configurations=version-1]{siunitx} % newer version

\theorembodyfont{\upshape}
\theoremheaderfont{\scshape}
\theorempostheader{:}
\theoremsep{\newline}

\jmlrvolume{tbd}
\jmlryear{2025}
\jmlrworkshop{International Conference on Computational Optimization}

\title[Quantum-Inspired Episode Selection for Monte Carlo Reinforcement Learning]{Quantum-Inspired Episode Selection for Monte Carlo Reinforcement Learning via QUBO Optimization}

\author{%
  \Name{Hadi Salloum} \Email{h.salloum@innopolis.ru}\\
  \addr Phystech School of Applied Mathematics and Computer Science, MIPT, Russia\\
  Research Center for Artificial Intelligence, Innopolis University, Russia\\
  Q Deep, Innopolis, Russia
  \AND
  \Name{Ali Jnadi} \Email{a.jnadi@innopolis.university}\\
  \addr Phystech School of Applied Mathematics and Computer Science, MIPT, Russia\\
  Research Center for Artificial Intelligence, Innopolis University, Russia\\
  Q Deep, Innopolis, Russia
  \AND
  \Name{Yaroslav Kholodov} \Email{y.kholodov@innopolis.ru}\\
  \addr Phystech School of Applied Mathematics and Computer Science, MIPT, Russia\\
  Innopolis, Russia\\
  Laboratory of Quantum Computing, Innopolis University, Russia
  \AND
  \Name{Alexander Gasnikov} \Email{gasnikov@yandex.ru}\\
  \addr Phystech School of Applied Mathematics and Computer Science, MIPT, Russia\\
  Research Center for Artificial Intelligence, Innopolis University, Russia
}
\begin{document}

\maketitle

\begin{abstract}
    Monte Carlo (MC) reinforcement learning suffers from high sample complexity, especially in environments with sparse rewards, large state spaces, and correlated trajectories. We address these limitations by reformulating episode selection as a Quadratic Unconstrained Binary Optimization (QUBO) problem and solving it with quantum-inspired samplers. Our method, MC+QUBO, integrates a combinatorial filtering step into standard MC policy evaluation: from each batch of trajectories, we select a subset that maximizes cumulative reward while promoting state-space coverage. This selection is encoded as a QUBO, where linear terms favor high-reward episodes and quadratic terms penalize redundancy. We explore both Simulated Quantum Annealing (SQA) and Simulated Bifurcation (SB) as black-box solvers within this framework. Experiments in a finite-horizon GridWorld demonstrate that MC+QUBO outperforms vanilla MC in convergence speed and final policy quality, highlighting the potential of quantum-inspired optimization as a decision-making subroutine in reinforcement learning.
\end{abstract}

\section{Introduction}

%minimizing \(H_{\mathrm{Ising}}\) is equivalent to maximizing a weighted cut and therefore captures NP-hard combinatorial structure:

%\begin{theorem}[Ising--MAX-CUT equivalence]
%For \(J_{ij}=w_{ij}\ge 0\), minimizing \(H_{\mathrm{Ising}}\) is equivalent (up to an additive constant) to maximizing the cut weight
%\begin{equation}
%\max_{\mathbf{s}\in\{\pm1\}^n} \sum_{(i,j)\in E} w_{ij}\frac{1 - s_i s_j}{2}.
%\end{equation}
%\end{theorem}
The Ising model \cite{cipra1987introduction}, originating in statistical mechanics, provides a fundamental framework for combinatorial optimization \cite{mohseni2022ising, bashar2023designing}. Defined on an undirected graph \(G=(V,E)\) with \(|V|=n\) vertices, its Hamiltonian for spin configurations \(\mathbf{s} \in \{\pm 1\}^n\) is  
\begin{equation}
H_{\mathrm{Ising}}(\mathbf{s}) = -\sum_{(i,j) \in E} J_{ij} s_i s_j - \sum_{i \in V} h_i s_i,
\label{eq:ising_hamiltonian}
\end{equation}
where \(J_{ij}\in\mathbb{R}\) represents pairwise couplings and \(h_i\in\mathbb{R}\) denotes local fields. For \(J_{ij}\ge 0\), the couplings are ferromagnetic (favoring aligned spins).

The Quadratic Unconstrained Binary Optimization (QUBO) formulation provides an equivalent representation using binary variables:
\begin{equation}
H_{\mathrm{QUBO}}(\mathbf{x}) = \mathbf{x}^\top Q \mathbf{x} + \mathbf{q}^\top \mathbf{x},
\qquad \mathbf{x}\in\{0,1\}^n,
\label{eq:qubo}
\end{equation}
where \(Q\in\mathbb{R}^{n\times n}\) is symmetric and may include diagonal terms encoding linear biases \cite{salloum2025mini, salloum2024quantum, salloum2024enhancing, salloum2025performance}.

The link between Ising spins and binary variables is given by:
\begin{equation}
s_i = 2x_i - 1.
\label{eq:spin_binary_map}
\end{equation}
Substituting \(s_i s_j = 4x_i x_j - 2x_i - 2x_j + 1\) into \eqref{eq:ising_hamiltonian} and collecting terms yields:
\begin{align}
H_{\mathrm{Ising}}(\mathbf{s}) &= 
\underbrace{\Big(-\sum_{i<j} J_{ij} + \sum_i h_i\Big)}_{C}
+ \sum_i \Big(2\sum_{j\ne i} J_{ij} - 2h_i\Big) x_i
+ \sum_{i<j} (-4J_{ij})\, x_i x_j,
\label{eq:ising_to_qubo}
\end{align}
so that a QUBO representation is obtained with:
\[
Q_{ij} = -4 J_{ij}\quad(i<j),\qquad
q_i = 2\sum_{j\ne i} J_{ij} - 2h_i,
\]
and an additive constant \(C\) (irrelevant for optimization).  
\emph{Remark:} different sign conventions for \(H_{\mathrm{Ising}}\) and \(H_{\mathrm{QUBO}}\) exist; the above is exact for the chosen convention. We consider the minimization form:
\[
\textsc{Min-QUBO:}\quad
\min_{x\in\{0,1\}^n} F(x) = x^\top Q x + q^\top x.
\]
This is \textbf{NP-hard}: solving it exactly in polynomial time would imply \(\mathrm{P}=\mathrm{NP}\).  
The related decision problem is:
\[
\textsc{QUBO-Decision:}\quad
\text{Given }Q,q,T,\ \exists x\in\{0,1\}^n\ \text{with }F(x) \le T?
\]
This is NP-complete. NP-hardness of \textsc{Min-QUBO} follows since an optimizer for all instances also solves the decision form \cite{salloumquantum, salloum2024quantuma}.

\paragraph{Proof sketch (reduction from Partition).}  
Given positive integers \(a_1,\dots,a_n\) and \(K=\frac12\sum_i a_i\), define:
\[
f(x) = \bigg(\sum_{i=1}^n a_i x_i - K\bigg)^2
      = x^\top (\mathbf{a}\mathbf{a}^\top) x - 2K\,\mathbf{a}^\top x + K^2.
\]
Set \(Q = \mathbf{a}\mathbf{a}^\top\) (rank 1), \(q = -2K\mathbf{a}\), constant \(C = K^2\).  
Then \(\min f(x) = 0\) iff there is a subset summing to \(K\) --- i.e., a Partition solution \cite{}.

\paragraph{Approximability and special cases.}
\begin{itemize}
  \item Some subclasses (e.g., MAX-CUT with nonnegative weights) admit constant-factor approximations (Goemans--Williamson $\approx 0.878$) \cite{goemans1995improved}.
  \item For arbitrary signed weights, uniform guarantees are generally impossible without structural restrictions.
  \item Special cases (e.g., positive semidefinite \(Q\) with sparse structure) can be solved efficiently.
\end{itemize}

\paragraph{Practical note.}  
Because exact solution is often infeasible for large \(n\), practical work uses:
\begin{itemize}
  \item \textbf{Exact methods:} integer programming, branch-and-bound for small/medium instances.
  \item \textbf{Relaxations:} SDP, spectral methods for bounds plus rounding heuristics.
  \item \textbf{Heuristics:} physics-inspired or metaheuristic approaches for large-scale best-effort solutions.
\end{itemize}

Two quantum-inspired approaches to QUBO/Ising are \emph{Simulated Quantum Annealing} (SQA) \cite{crosson2016simulated} and \emph{Simulated Bifurcation} (SB) \cite{goto2019combinatorial}. In this work, our aim is to integrate such solvers into \emph{reinforcement learning} (RL) agents, enabling adaptive strategies that leverage physics-inspired optimization within decision-making frameworks.  
To the best of our knowledge, only a few works have explored RL--solver integration, and existing studies are limited in scope and quality. 

\begin{figure}
    \centering
    \includegraphics[width=0.95\linewidth]{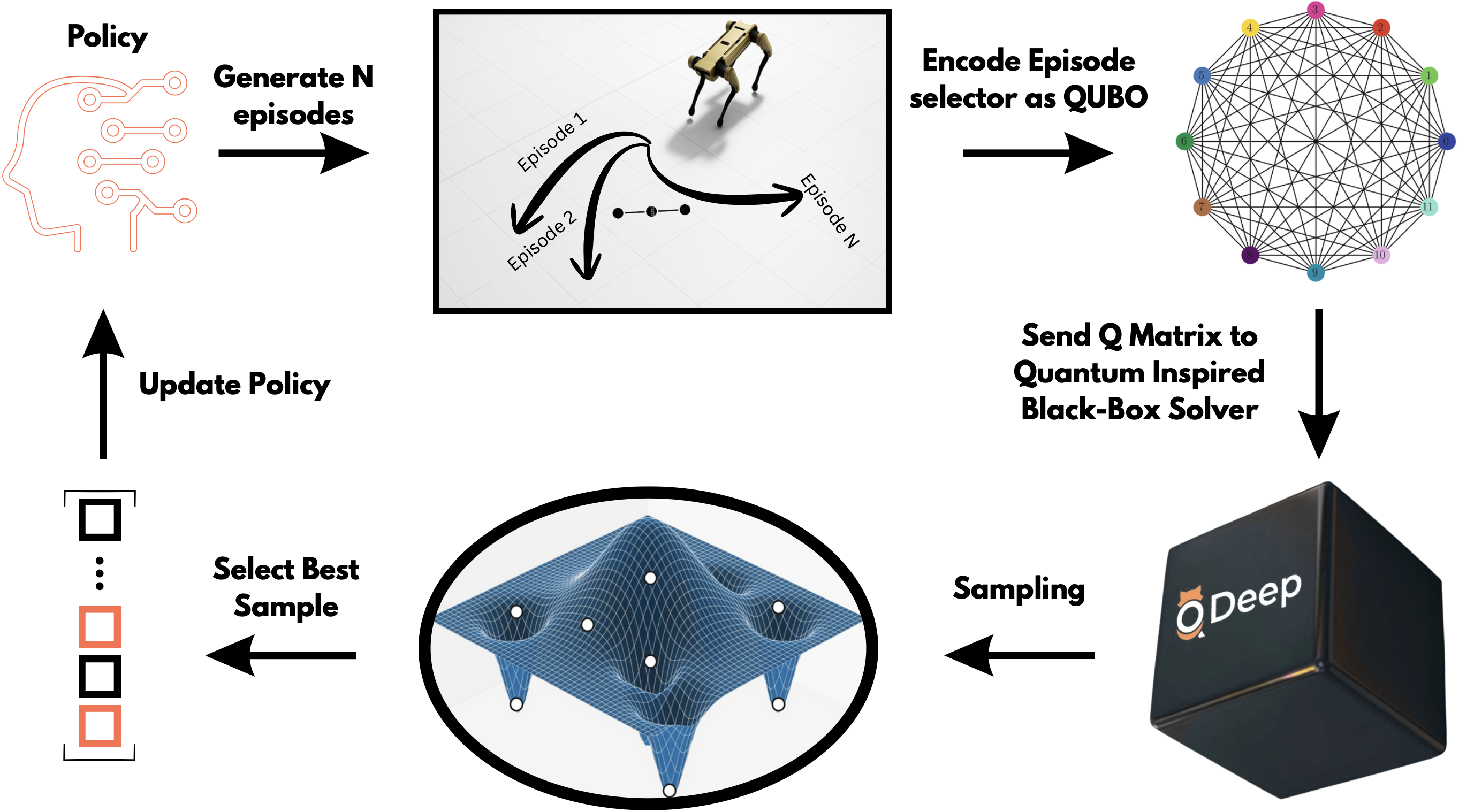}
    \caption{Iterative Quantum-Assisted Reinforcement Learning — The policy generates multiple episodes, which are encoded as a QUBO and solved by a quantum black-box solver. The best sample guides the policy update, and the process repeats iteratively to converge toward the optimal policy.}
    \label{fig:work}
\end{figure}

We therefore begin with a \emph{value-function-approximation} RL setting and choose \emph{Monte Carlo} methods as our starting point, as they provide a simple, unbiased baseline for exploring this integration. SQA mimics quantum annealing (please refer to \cite{kadowaki1998quantum, morita2008mathematical} understand the concept of quantum annealing and its mathematical foundation) using classical stochastic updates (e.g., path-integral Monte Carlo) rather than real quantum hardware.  
It interpolates between a driver Hamiltonian \(H_0\) and the problem Hamiltonian \(H_P\):

\begin{equation}
    H_{\mathrm{SQA}}(t) = A(t) H_0 + B(t) H_P,\quad t\in[0,T],
\end{equation}

with \(A(0)=1,\ B(0)=0\) and \(A(T)=0,\ B(T)=1\).  
The simulation samples trajectories from an effective quantum partition function, using a discretized ``imaginary time'' dimension to represent quantum fluctuations.  
These fluctuations allow the simulated system to perform \emph{quantum-like tunneling} between separated minima, a mechanism that can escape traps more efficiently than purely thermal hops in simulated annealing. The performance of SQA depends on the annealing schedule \((A(t),B(t))\), the number of Trotter slices (which controls the resolution of quantum fluctuations), and the Monte Carlo update strategy.  
Proper tuning balances exploration (large quantum fluctuations) and exploitation (classical convergence) as \(t\) approaches \(T\).  
While SQA is not a quantum algorithm in the strict sense, it inherits many behaviors of quantum annealing and is attractive as a software-based, hardware-independent tool. In the other hand, SB is a classical dynamical system approach that drives continuous variables toward binary states via bifurcation phenomena:
\begin{equation}    
\dot{x}_i = y_i,\quad
\dot{y}_i = -(\mu(t)-\lambda)x_i + \sum_j J_{ij}x_j - \gamma y_i.
\end{equation}

Here \(\mu(t)\) ramps through a critical value \(\mu_c = \lambda + \rho(J)\) (spectral radius of \(J\)), triggering a symmetry-breaking transition that favors low-energy configurations.  
Parameter choices for \(\gamma\), \(\lambda\), and the ramp schedule crucially influence solution quality, and empirical tuning is often necessary.

\paragraph{Analogy and distinction.}  
Both SQA and SB use gradual parameter changes to steer the system toward good solutions:
\begin{itemize}
  \item SQA uses \emph{stochastic sampling} to emulate quantum transitions, with tunable quantum fluctuations.
  \item SB uses \emph{deterministic dynamics} with bifurcation-induced binary state formation.
\end{itemize}
We treat both as black-box oracles:
\begin{equation}
\mathbb{P}\big(H_{\mathrm{QUBO}}(\mathbf{x}^*) \le \min_{\mathbf{x}} H_{\mathrm{QUBO}}(\mathbf{x}) + \delta\big) \ge 1 - \eta,
\end{equation}

allowing their integration into RL agents.  
Our starting point is to use quantum inspired computing as action-selection subroutines within Monte Carlo-based RL as shown in Figure \ref{fig:work}, building toward more sophisticated architectures. We will use the Qonquester Cloud platform by Q Deep to implement and run these quantum-inspired components \cite{qdeepnet}.

\section{Sample Efficiency Limitations in Monte Carlo Reinforcement Learning}

A central limitation of most modern stochastic optimization methods is their reliance on the assumption that data samples are independently and identically distributed. In reinforcement learning, this assumption is inherently violated, as samples are temporally correlated through the underlying Markov decision process (MDP). Furthermore, the effective application of many algorithms presupposes knowledge of the MDP’s mixing time or asymptotic behavior. For MDPs with high-dimensional state spaces or sparse reward structures, the mixing time is often intractable to estimate \cite{wolfer20} —if not entirely unknown—thereby restricting the applicability of such methods in practice. Monte Carlo (MC) methods estimate the value function based on the returns obtained from visiting a given state. However, these estimates are inherently tied to the policy under which the trajectories are generated, and that policy is not guaranteed to be optimal. Consequently, the value estimates may fail to converge to the optimal solution \cite{LIU2021109693, Winnicki2023}. A common remedy is to employ an exploration–exploitation strategy \cite{Sutton1998}, in which the agent occasionally takes random actions (exploration) to discover new states, while otherwise selecting actions according to the current value function (exploitation). While this approach mitigates policy bias, it can also slow convergence and introduce instability in the learning process. For state $s$, the MC estimator after $N$ episodes is:

\begin{equation}
\hat{V}(s) = \frac{1}{N_s} \sum_{k=1}^{N} \mathbb{I}_k(s) G_k(s)
\end{equation}

where $G_k(s)$ is the cumulative return from $s$ in episode $k$, $N_s = \sum_k \mathbb{I}_k(s)$ counts visits to $s$, and $\mathbb{I}_k(s)$ indicates visitation. This approach demands large $N$ due to:

\begin{enumerate}
\item \textit{Variance Propagation}: Returns exhibit variance compounding:
\begin{equation}
\text{Var}[G_k(s)] \geq \frac{\sigma_R^2}{1-\gamma^2}
\end{equation}
where $\sigma_R^2$ is reward variance and $\gamma$ the discount factor. Reducing estimation error below $\epsilon$ requires $N_s > \mathcal{O}(\epsilon^{-2})$ episodes.

\item \textit{Suboptimal Trajectory Dilution}: The estimator decomposes as:
\begin{equation}
\hat{V}(s) = \alpha \mathbb{E}[G|\mathcal{T}{\text{opt}}] + (1-\alpha) \mathbb{E}[G|\mathcal{T}{\text{sub}}]
\end{equation}
where $\alpha = |\mathcal{T}_{\text{opt}}|/N_s$. Convergence requires $\alpha > 0$, necessitating exponential growth in $N$ for sparse-reward domains.

\item \textit{Exploration-Computation Duality}: $\epsilon$-optimal policy convergence requires:
\begin{equation}
N_s \geq \frac{\log(|\mathcal{S}|/\delta)}{2\epsilon^2(1-\gamma)^2}
\end{equation}
by PAC analysis, becoming prohibitive for large $|\mathcal{S}|$.
\end{enumerate}

Theoretically, selecting an optimal subset $\mathcal{E}^* \subset \mathcal{E}$ of $m$ episodes ($m \ll N$) could overcome these limitations by solving:

\begin{equation}
\min_{\mathcal{E}^* \subset \mathcal{E}} \left| \hat{V}_{\mathcal{E}^*}(s) - V^{\pi}(s) \right|_2 \quad \text{s.t.} \quad |\mathcal{E}^*| = m
\end{equation}

This is isomorphic to feature selection where episodes constitute basis functions. However, this optimization exhibits:

\begin{enumerate}
\item \textit{Combinatorial Explosion}: The solution space has $\binom{N}{m}$ configurations, becoming infeasible for $N > 50$.

\item \textit{Nonlinear Coupling}: Estimation accuracy depends on trajectory interactions:
\begin{equation}
\Delta \hat{V} = \sum_{i \in \mathcal{E}^{*}} w_i G_i + \sum_{i \neq j \in \mathcal{E}^{*}} w_{ij} G_i G_j + \mathcal{O}(G^3)
\end{equation}

\item \textit{NP-Hardness}: The problem reduces to maximum coverage:
\begin{equation}
\max_{\mathbf{x}} \sum_{s \in \mathcal{S}} \min\left(1, \sum_i x_i \mathbb{I}_i(s)\right) \quad \text{s.t.} \quad \sum_i x_i = m, x_i \in {0,1}
\end{equation}
which is NP-hard by reduction from set cover.
\end{enumerate}

As established in Section 2, quantum computing and its inspired versions provide a viable approach for such combinatorial problems. The selection problem admits a natural QUBO formulation:

\begin{equation}
\mathbf{x}^* = \argmin_{\mathbf{x} \in {0,1}^N} \left( -\mathbf{c}^T\mathbf{x} + \mathbf{x}^T\mathbf{Q}\mathbf{x} \right)
\end{equation}

where $\mathbf{c}$ encodes episode rewards and $\mathbf{Q}$ controls trajectory similarity penalties. Quantum annealing navigates this energy landscape with expected complexity $\mathcal{O}(\exp(\sqrt{N}))$ versus classical $\mathcal{O}(\exp(N))$ for sparse QUBOs. This paradigm transforms MC-RL from statistical averaging to combinatorial optimization. We now formalize the QUBO representation for quantum processing.

\section{Monte Carlo Policy Learning with QUBO-Based Episode Selection}

We study policy learning in a stochastic gridworld where an agent collects batches of episodic trajectories and updates an action-value table \(Q(s,a)\) via first-visit Monte Carlo (MC). Two procedures are considered: (i) a baseline \emph{Vanilla MC} that uses all sampled episodes for each update (Algorithm~\ref{alg:mc}), and (ii) \emph{MC+QUBO}, which inserts a combinatorial filtering step between sampling and policy update (Algorithm~\ref{alg:mc-qubo}). In MC+QUBO, from \(n\) sampled episodes \(\{\tau_i\}_{i=1}^n\) (each with visited-state set \(S_i\) and cumulative reward \(R_i=\sum_t r_t(\tau_i)\)) we solve a Quadratic Unconstrained Binary Optimization (QUBO) to select a compact subset that balances high return and coverage (diversity) prior to performing MC updates.

The Vanilla MC baseline is given as Algorithm~\ref{alg:mc}. The MC+QUBO pipeline (Algorithm~\ref{alg:mc-qubo}) differs only in that, after sampling a batch of episodes, it canonicalizes episodes (optional), computes episode rewards \(R_i\) and pairwise similarities \(w_{ij}\in[0,1]\) (e.g. Jaccard similarity \(w_{ij}=|S_i\cap S_j|/|S_i\cup S_j|\)), and then formulates and dispatches the following QUBO to a quantum-inspired black-box sampler:

\begin{equation}\label{eq:qubo_main}
\min_{x\in\{0,1\}^n}
\; 
\sum_{i=1}^n \big(-\alpha\,R_i\big)\,x_i
\;+\;
\sum_{i=1}^n\sum_{j=i+1}^n \gamma\,w_{ij}\,x_i x_j,
\end{equation}
where \(x_i=1\) selects episode \(\tau_i\), \(\alpha>0\) is a reward-scaling multiplier, and \(\gamma\ge 0\) weights the similarity penalty. In solver form the QUBO is \(x^\top Q_{\mathrm{qubo}} x\) with \((Q_{\mathrm{qubo}})_{ii}=-\alpha R_i\) and \((Q_{\mathrm{qubo}})_{ij}=\tfrac{1}{2}\gamma w_{ij}\) for \(i\neq j\) (adapt to solver conventions) \footnote{Normalize $R_i$ and $w_{ij}$, or tune $\alpha,\gamma$ so coefficient magnitudes are comparable. If a fixed cardinality $k$ is required, add a soft penalty $\lambda\left(\sum_i x_i - k\right)^2$. Allocate sufficient sampling budget to the black-box solver to reliably observe low-energy solutions, and consider repeating the QUBO solve and aggregating best samples for robustness.}.

Equation~\ref{eq:qubo_main} encodes a trade-off: the linear term \(-\alpha R_i x_i\) favors high-return episodes (minimization reduces objective when selecting large \(R_i\)), while the quadratic term \(\gamma w_{ij} x_i x_j\) penalizes selecting similar episodes and thus promotes diversity. The reward multiplier \(\alpha\) is intentionally used to enlarge the energy gap between high- and low-quality episodes so that similarity penalties do not dominate and lead the solver to select low-reward but superficially diverse sets. A practical heuristic for \(\alpha\) is
\[
\alpha\cdot\mathbb{E}[R] \;\gtrsim\; \gamma\cdot (k-1)\cdot\mathbb{E}[w],
\]
where \(k\) is an expected selection size; this ensures the linear benefit of a representative good episode typically outweighs the incremental quadratic penalty of adding it to an already-selected set.

The QUBO is submitted to a quantum-inspired black-box sampler which returns a pool of candidate samples \(\{x^{(s)}\}\) together with energies (objective values). Standard practice is to request many samples, sort by energy, and take the best-observed sample \(x^\star=\arg\min_s \mathrm{energy}(x^{(s)})\) — or to aggregate the top few low-energy samples (e.g., majority vote per bit) to increase robustness. The indices \(\mathcal{I}=\{i:x^\star_i=1\}\) define the episode subset used for MC updates. This sampling-and-selection workflow is reflected in Algorithm~\ref{alg:mc-qubo} (lines that call the sampler and choose the best sample).

Algorithms~\ref{alg:mc} and~\ref{alg:mc-qubo} are presented below in pseudocode form. Algorithm~\ref{alg:mc} is the all-episode baseline; Algorithm~\ref{alg:mc-qubo} contains the full QUBO construction, dispatch to a quantum-inspired sampler, sampling/selection step (best sample chosen), and subsequent first-visit MC updates performed only on the selected episodes.

\begin{algorithm2e}
\caption{Vanilla Monte Carlo (use all episodes)}
\label{alg:mc}
\KwIn{number of batches $N$, batch size $m$, environment, slip probability}
\KwOut{policy $\pi$, action-value estimates $Q(s,a)$}

Initialize policy $\pi$\;
Initialize $Q(s,a) \gets 0$ and empty returns history for all $(s,a)$\;

\For{$\text{batch} \gets 1$ \KwTo $N$}{
    Generate $m$ episodes $\{\tau_1,\dots,\tau_m\}$ by following $\pi$ (with slip)\;
    
    \For{each episode $\tau$ in the batch}{
        \For{each first-visit $(s,a)$ in $\tau$}{
            Compute return $G$ (sum of future rewards from this visit onward)\;
            Append $G$ to $\mathrm{returns}(s,a)$\;
            Update $Q(s,a) \gets \mathrm{mean}(\mathrm{returns}(s,a))$\;
        }
    }
    
    Update policy $\pi(s) \gets \arg\max_a Q(s,a)$ for all states $s$\;
}
\end{algorithm2e}

\begin{algorithm2e}
\caption{Monte Carlo + QUBO Episode Selection (sampling + best-sample selection)}
\label{alg:mc-qubo}
\KwIn{number of batches $N$, batch size $m$, reward weight $\alpha$, similarity weight $\gamma$, QUBO sampler, optional target size $k$}
\KwOut{policy $\pi$, action-value estimates $Q(s,a)$}

Initialize policy $\pi$\;
Initialize $Q(s,a) \gets 0$ and empty returns history for all $(s,a)$\;

\For{$\text{batch} \gets 1$ \KwTo $N$}{
    Generate $m$ episodes $\{\tau_1,\dots,\tau_m\}$ by following $\pi$\;
    Optionally canonicalize each episode to first-visit representation\;
    
    \tcp{Compute rewards and visited sets}
    \For{$i \gets 1$ \KwTo $m$}{
        $R_i \gets$ total reward of episode $\tau_i$\;
        $S_i \gets$ set of visited state--action pairs (or states) in $\tau_i$\;
    }
    
    \tcp{Compute pairwise similarities (e.g. Jaccard)}
    \For{$i \gets 1$ \KwTo $m-1$}{
        \For{$j \gets i+1$ \KwTo $m$}{
            $w_{ij} \gets \text{similarity}(S_i, S_j) \in [0,1]$\;
        }
    }
    
    \tcp{Form QUBO coefficients}
    \For{$i \gets 1$ \KwTo $m$}{
        $h_i \gets -\alpha R_i$\;
    }
    \For{$i \gets 1$ \KwTo $m-1$}{
        \For{$j \gets i+1$ \KwTo $m$}{
            $J_{ij} \gets \gamma w_{ij}$\;
        }
    }
    
    Submit QUBO $(h, J)$ to quantum-inspired black-box sampler; request many samples\;
    Receive sample pool $\{x^{(s)}\}$ with energies\;
    Choose best sample $x^\star \gets \arg\min_s \mathrm{energy}(x^{(s)})$\;
    Optionally post-process top samples (vote / ensemble) to form a robust $x^\star$\;
    
    $\mathcal{I} \gets \{ i \mid x^\star_i = 1 \}$\;
    Selected episodes: $\{\tau_i\}_{i \in \mathcal{I}}$\;
    
    \For{each selected episode $\tau_i$}{
        \For{each first-visit $(s,a)$ in $\tau_i$}{
            Compute return $G$\;
            Append $G$ to $\mathrm{returns}(s,a)$\;
            Update $Q(s,a) \gets \mathrm{mean}(\mathrm{returns}(s,a))$\;
        }
    }
    
    Update policy $\pi(s) \gets \arg\max_a Q(s,a)$ for all states $s$\;
}
\end{algorithm2e}

\section{Experiments and Results}
We evaluated our proposed MC+QUBO episode selection method on finite-horizon GridWorld environments of sizes \(\{3\times 3,\, 5\times 5,\, 8\times 8,\, 10\times 10,\, 15\times 15,\, 20\times 20\}\), comparing it against the baseline vanilla MC algorithm. In each setting, the agent was trained over multiple batches of episodes, with performance measured in terms of convergence speed, policy quality, and stability.

Across all tested grid sizes, MC+QUBO consistently converged in fewer batches than vanilla MC, with the improvement becoming more pronounced in larger environments (\(\geq 10\times 10\)), where sparse rewards and the exponential growth of the state space typically hinder policy evaluation. By integrating QUBO-based episode selection into the MC framework, the algorithm focuses learning on \emph{informative and diverse trajectories}, avoiding redundant episodes that offer little new information for value estimation. This selection is formulated as a QUBO problem, solved using Quantum Inspired Black Box Solver, delivered solutions within practical computational budgets.

A key design choice in our implementation is not directly optimizing for reward during episode selection. Although the QUBO formulation includes a reward term, our main experiments set its influence close to zero, prioritizing state-space coverage instead. This mitigates the risk of overfitting to a narrow set of high-reward episodes, which could bias policy updates and reduce exploration. Preliminary tests confirmed that reward-driven selection often leads to faster but less stable convergence, whereas our reward-independent formulation yielded \emph{more balanced learning} and better generalization.

\begin{figure}[htbp]
    \centering

    % First row
    \begin{minipage}[b]{0.485\textwidth}
        \centering
        \includegraphics[width=\textwidth]{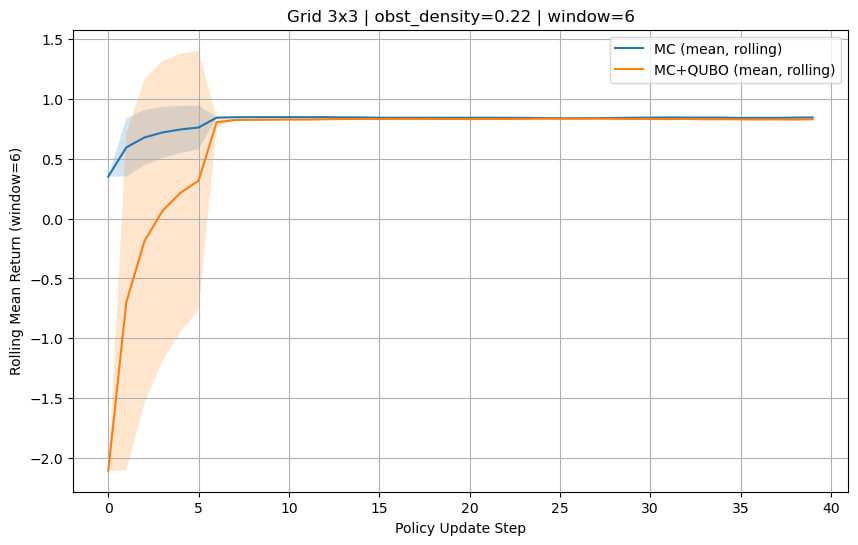}
    \end{minipage}
    \hfill
    \begin{minipage}[b]{0.485\textwidth}
        \centering
        \includegraphics[width=\textwidth]{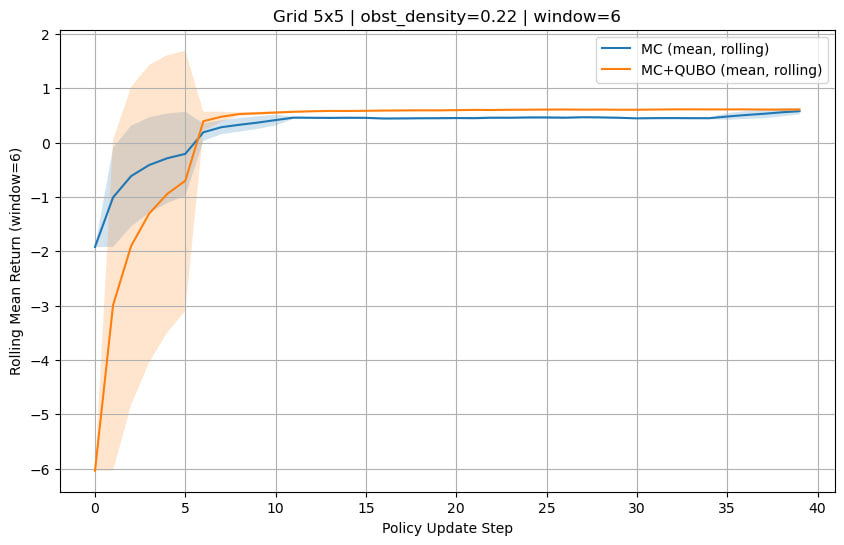}
    \end{minipage}
    \vspace{1em}

    \begin{minipage}[b]{0.485\textwidth}
        \centering
        \includegraphics[width=\textwidth]{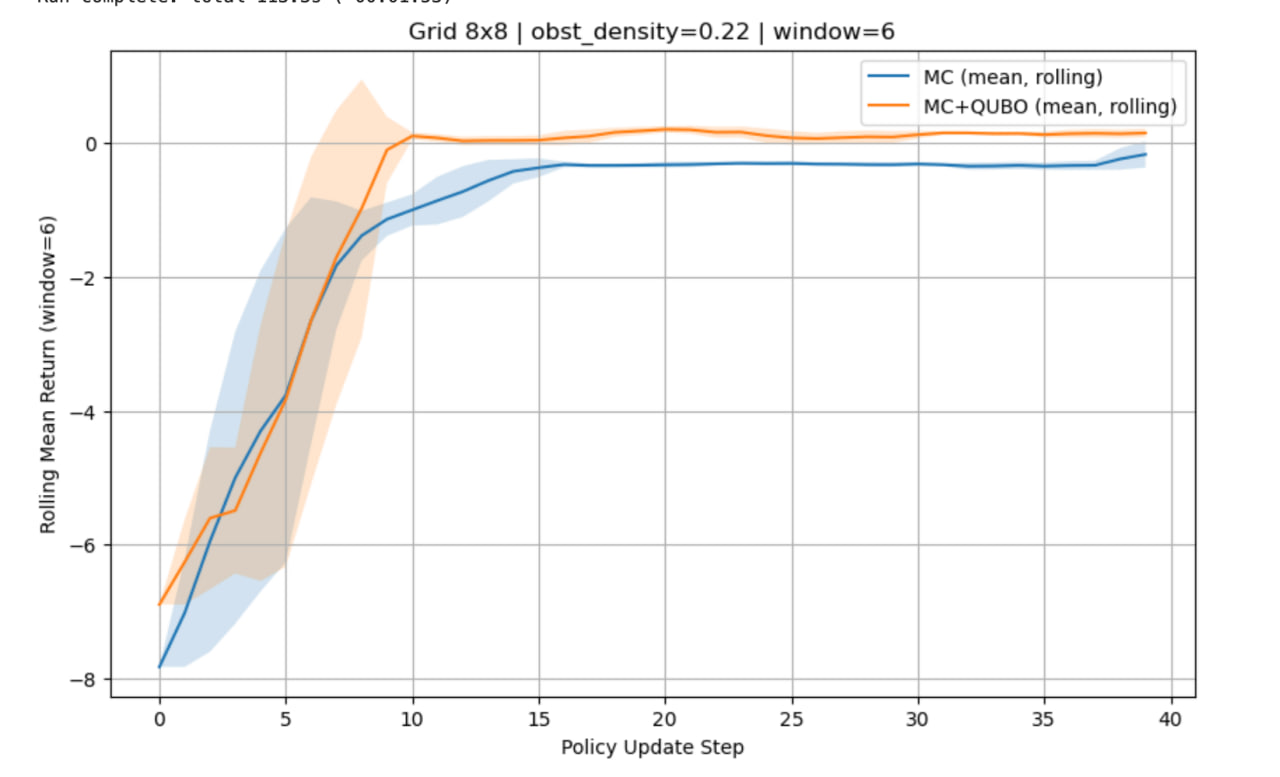}
    \end{minipage}
    \hfill
    \begin{minipage}[b]{0.485\textwidth}
        \centering
        \includegraphics[width=\textwidth]{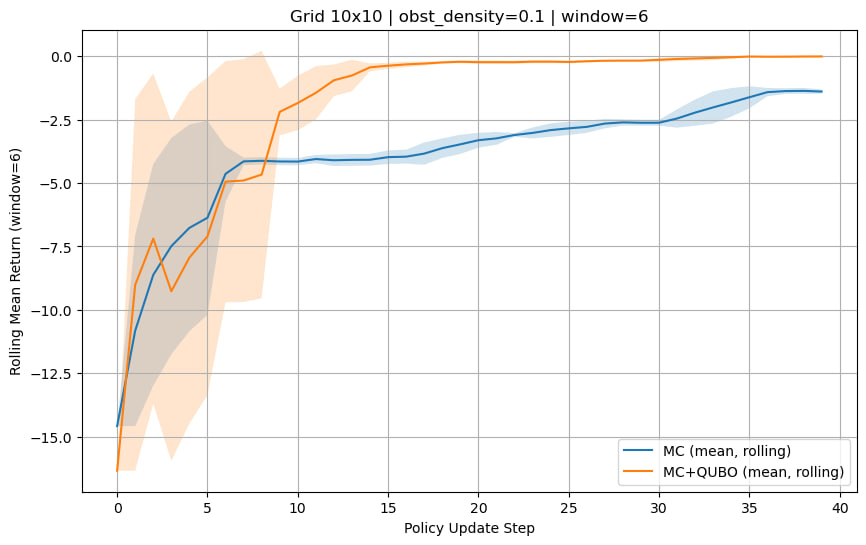}
    \end{minipage}
    \vfill
    \begin{minipage}[b]{0.485\textwidth}
        \centering
        \includegraphics[width=\textwidth]{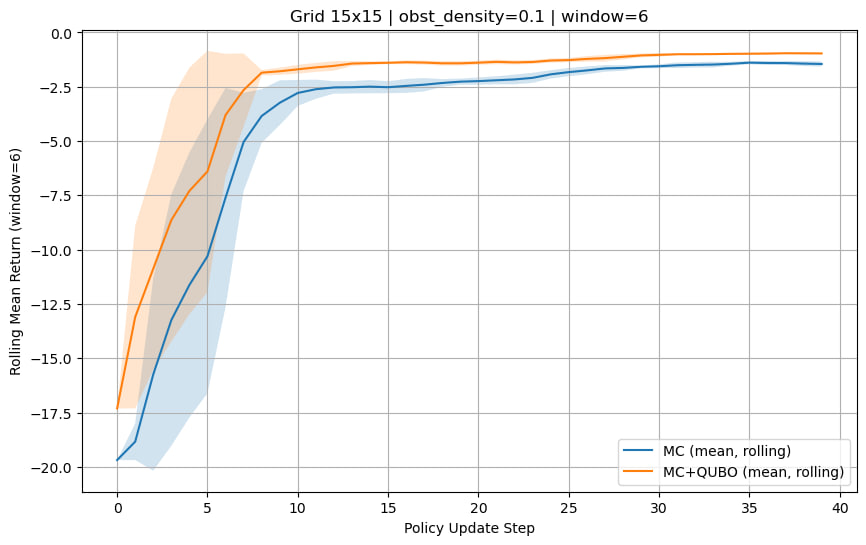}
    \end{minipage}
    \hfill
    \begin{minipage}[b]{0.485\textwidth}
        \centering
        \includegraphics[width=\textwidth]{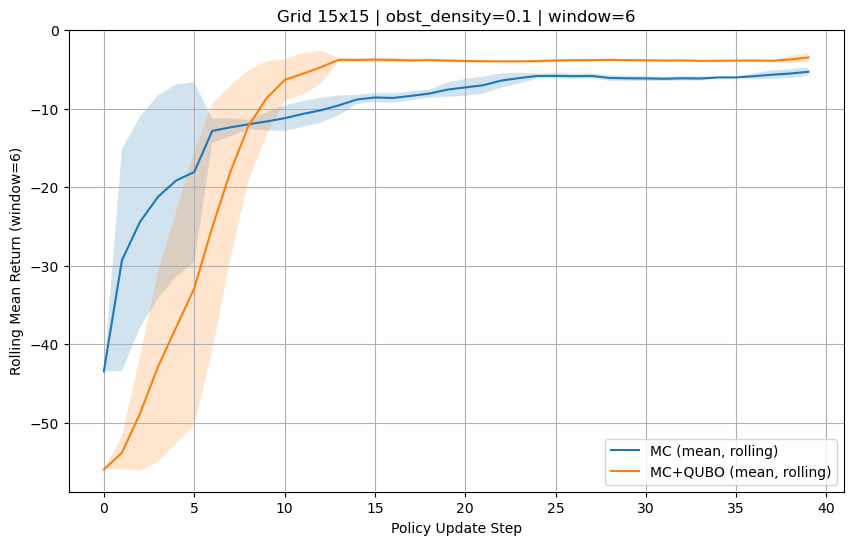}
    \end{minipage}

  \caption{Rolling mean returns (with a window size of 6) over policy update steps for Monte Carlo (MC) and MC combined with Quadratic Unconstrained Binary Optimization (QUBO) methods in gridworld environments. The top row displays results for smaller grids (3×3, 5×5, 8×8) with an obstacle density of 0.22, while the bottom row shows larger grids (10×10, 15×15) with an obstacle density of 0.1 and (20×20) with an obstacle density of 0.01. Blue lines represent the MC method, orange lines represent the MC+QUBO method, and shaded regions indicate variability in the returns.}
  \label{fig:all_six}
\end{figure}

The resulting policies, illustrated in Figure~\ref{fig:all_six}, achieved \textbf{higher average returns} than those obtained with vanilla MC across all environments. The margin of improvement was greatest in larger grids, where MC+QUBO maintained exploratory diversity while accelerating convergence. Furthermore, the scalability of our QUBO approach demonstrates that quantum-inspired optimization can be feasibly integrated as a decision-making subroutine in reinforcement learning.

% ===== DISCUSSION AND REMARKS SECTION =====

\section{Discussion and Remarks}

Our results demonstrate MC+QUBO's consistent superiority over vanilla Monte Carlo, particularly in larger grids ($\geq 10\times 10$). The quantum-inspired episode selection addresses key MC limitations:
\begin{itemize}
    \item \textbf{Sample redundancy} via quadratic similarity penalties
    \item \textbf{Sparse-reward dilution} through coverage prioritization
    \item \textbf{Variance propagation} by filtering correlated episodes
\end{itemize}

\textbf{Solver Implementation:} We used Qonquester Cloud to execute both SB and its thermal-assisted variants \cite{kanao2022simulated}, as well as SQA. Quantum-inspired solvers outperformed classical approaches (Tabu Search, Simulated Annealing), though these results are omitted for brevity \footnote{We have also conducted separate, unpublished benchmarking of quantum-inspired solvers (SB, SQA) against classical heuristics (Tabu Search, Simulated Annealing) on standalone QUBO problems of 100–2000 variables. Preliminary results indicated that quantum-inspired solvers often produce lower-energy solutions for our episode-selection QUBOs. While these findings support the solver choice in this work, they are not part of the formal evaluation here and will be detailed in a forthcoming publication.}.

\textbf{Computation:} QUBO solving latency ($\sim$0.5-2s/batch) was dominated by \textbf{cloud communication} (matrix transfer). Actual solver time was negligible ($\sim$10-100ms) for $n \leq 200$.

\section{Conclusion}
Our results confirm that QUBO-based episode selection offers a practical and effective enhancement to Monte Carlo reinforcement learning, particularly in environments with sparse rewards, large state spaces, or correlated trajectories. Beyond GridWorld, the framework could be extended to continuous control problems, hierarchical RL, or multi-agent systems, with future work exploring dynamic tuning of selection weights, hybrid criteria, and deployment on real quantum hardware. By bridging RL with combinatorial optimization, this study opens the door to a new class of algorithms where quantum and quantum-inspired methods actively guide the learning process.

\acks{The study was supported by the Ministry of Economic Development of the Russian Federation (agreement No. 139-10-2025-034 dd. 19.06.2025, IGK 000000C313925P4D0002)}

\bibliography{sample}

\end{document}